\documentclass[letterpaper]{article}
\usepackage{aaai}
\usepackage{times}
\usepackage{helvet}
\usepackage{courier}
\usepackage{graphicx}
\usepackage{subfigure} 

\pdfpageattr {/Group << /S /Transparency /I true /CS /DeviceRGB>>}

\newcommand{\citenp}[1]{\citeauthor{#1}}
\newcommand{\citep}[1]{\cite{#1}}
\newcommand{\citet}[1]{\citeauthor{#1}~\shortcite{#1}}

\def\plotheight{60pt}

\usepackage[usenames,dvipsnames]{xcolor}
\usepackage{amssymb,amsfonts,amsbsy,amsmath,amsthm}
\theoremstyle{plain}
\newtheorem{thm}{Theorem}
\newtheorem{theorem}[thm]{Theorem}

\theoremstyle{definition}

\newtheorem{defi}{Definition}
\newtheorem{definition}[defi]{Definition}

\newcommand{\bm}[1]{\mathbf{#1}}
\DeclareMathAlphabet{\mathbfsf}{\encodingdefault}{\sfdefault}{bx}{n}

\def\P{P}

\def\xs{\bm{x}}

\def\Xs{\bm{X}}

\def\xl{\mathsf{x}}

\def\ys{\bm{y}}

\def\yl{\mathsf{y}}

\newcommand{\mln}[1]{#1}

\newcommand{\dom}{\mathbfsf{D}}

\newcommand{\has}{\mathtt{HasWord}}
\newcommand{\linked}{\mathtt{Link}}
\newcommand{\pageclass}{\mathtt{Page}}

\newcommand{\Course}{\mathsf{Course}}
\newcommand{\Faculty}{\mathsf{Faculty}}
\newcommand{\Hours}{\mathsf{Hours}}
\newcommand{\pagea}{\mathsf{A}}
\newcommand{\pageb}{\mathsf{B}}

\makeatletter 
\newcommand\mynobreakpar{\par\nobreak\@afterheading} 
\makeatother

\nocopyright

\begin{document}
%
\title{Lifted Probabilistic Inference for Asymmetric Graphical Models}

\author{
Guy Van den Broeck\\
Department of Computer Science\\KU Leuven, Belgium\\ \texttt{guy.vandenbroeck@cs.kuleuven.be}
\And 
Mathias Niepert\\
Computer Science and Engineering\\University of Washington, Seattle\\ \texttt{mniepert@cs.washington.edu}
}

\maketitle

\begin{abstract}
\begin{quote}
Lifted probabilistic inference algorithms have been successfully applied to a large number of symmetric graphical models. Unfortunately, the majority of real-world graphical models is asymmetric. This is even the case for relational representations when evidence is given. Therefore, more recent work in the community moved to making the models symmetric and then applying existing lifted inference algorithms. However, this approach has two shortcomings. First, all existing over-symmetric approximations require a relational representation such as Markov logic networks. Second, the induced symmetries often change the distribution significantly, making the computed probabilities highly biased. We present a framework for probabilistic sampling-based inference that only uses the induced approximate symmetries to propose steps in a Metropolis-Hastings style Markov chain. The framework, therefore, leads to improved probability estimates while remaining unbiased. Experiments demonstrate that the approach outperforms existing MCMC algorithms. 
\end{quote}
\end{abstract}

\section{Introduction}

Probabilistic graphical models are successfully used  in a wide range of applications. Inference in these models is intractable in general and, therefore, approximate algorithms are mostly applied. However, there are several probabilistic graphical models for which inference is tractable due to (conditional) independences and the resulting low treewidth~\cite{darwiche2009modeling,koller:2009}. Examples of the former class of models are chains and tree models. More recently, the AI community has uncovered additional statistical properties based on symmetries of the graphical model that render inference tractable~\cite{DBLP:conf/aaai/NiepertB14}. In the literature, approaches exploiting these symmetries are often referred to as lifted or symmetry-aware inference algorithms~\cite{Poole2003,Kersting:2012}. 

While lifted inference algorithms perform well for highly symmetric graphical models, they depend heavily on the presence of symmetries and perform worse for asymmetric models due to their computational overhead. This is especially unfortunate as numerous real-world graphical models are not symmetric. To bring the achievements of the lifted inference community to the mainstream of machine learning and uncertain reasoning it is crucial to explore ways to apply ideas from the lifted inference literature to inference problems in asymmetric graphical models.

Recent work has introduced methods to generate symmetric approximations of probabilistic models~\cite{conf/nips/BroeckD13,DBLP:conf/pkdd/VenugopalG14,singla2014approximate}. All of these approaches turn approximate symmetries, that is, symmetries that ``almost" hold in the probabilistic models, into perfect symmetries, and proceed to apply lifted inference algorithms to the symmetrized model. These approaches were shown to perform well but are also limited in a fundamental way. The introduction of artificial symmetries  results in marginal probabilities that are different from the ones of the original model. The per variable Kullback-Leibler divergence, a measure often used to assess the performance of approximate inference algorithms, might improve when these symmetries are induced but it is possible that the marginals the user actually cares about are highly biased. Of course, this is a potential problem in applications. For instance, consider a medical application where one queries the probability of diseases given symptoms. A symmetric approximation may perform well in terms of the KL divergence but might skew the probabilities of the most probable diseases to become equal. A major argument for graphical models is the need to detect subtle differences in the posterior, which becomes impossible when approximate symmetries skew the distribution.

To apply lifted inference to asymmetric graphical models we propose a completely novel approach. As in previous approaches, we compute a symmetric approximation of the original model but leverage the symmetrized model to compute a proposal distribution for a Metropolis-Hastings chain. The approach combines a base MCMC algorithm such as the Gibbs sampler with the Metropolis chain that performs jumps in the symmetric model. The novel framework allows us to utilize work on approximate symmetries such as color passing algorithms~\cite{DBLP:conf/aaai/KerstingMGG14} and low-rank Boolean matrix approximations~\cite{conf/nips/BroeckD13} while producing unbiased probability estimates. We identify properties of an approximate symmetry group that make it suitable for the novel lifted Metropolis-Hastings approach.

We conduct experiments where, for the first time, lifted inference is applied to graphical models with no exact symmetries and no color-passing symmetries, and where every random variable has distinct soft evidence. The framework, therefore, leads to improved probability estimates while remaining unbiased. Experiments demonstrate that the approach outperforms existing MCMC algorithms.

\section{Background}

We review some background on concepts and methods used throughout the remainder of the paper.

\subsection{Group Theory}

A group is an algebraic structure ($\mathfrak{G}, \circ$), where $\mathfrak{G}$ is a set closed under a binary associative operation $\circ$ with an identity element and a unique inverse for each element. We often write $\mathfrak{G}$ rather than  ($\mathfrak{G}, \circ$). A permutation group acting on a set $\Omega$ is a set of bijections  $\mathfrak{g} : \Omega \rightarrow \Omega$ that form a group.
Let $\Omega$ be a finite set and let $\mathfrak{G}$ be a permutation group acting on~$\Omega$. If $\alpha \in \Omega$ and $\mathfrak{g} \in \mathfrak{G}$ we write $\alpha^\mathfrak{g}$ to denote the image of $\alpha$ under $\mathfrak{g}$. 
A cycle $(\alpha_1\ \alpha_2\ ...\ \alpha_n)$
represents the permutation that maps $\alpha_1$ to $\alpha_2$, $\alpha_2$
to $\alpha_3$,..., and $\alpha_n$ to $\alpha_1$.
Every permutation can be written as a product of disjoint cycles. A generating set $R$ of a group is a subset of the group's elements such that every element of the group can be written as a product of finitely many elements of $R$ and their inverses.
 
We define a relation $\sim$ on $\Omega$ with $\alpha \sim \beta$ if and only if there is a permutation $\mathfrak{g} \in \mathfrak{G}$ such that $\alpha^\mathfrak{g} = \beta$. The relation partitions $\Omega$ into equivalence classes which we call orbits. We call this partition of $\Omega$ the orbit partition induced by $\mathfrak{G}$. We use the notation $\alpha^{\mathfrak{G}}$ to denote the orbit $\{\alpha^\mathfrak{g}\ |\ \mathfrak{g} \in \mathfrak{G}\}$ containing $\alpha$. 



\subsection{Symmetries of Graphical Models}

Symmetries of a set of random variables and graphical models have been formally defined in the lifted and symmetry-aware probabilistic inference literature with concepts from group theory~\cite{niepertorbits,hai2012automorphism}.

\begin{definition}
Let $\Xs$ be a set of discrete random variables with distribution $\pi$ and let $\Omega$ be the set of states (configurations) of $\Xs$. We say that a permutation group $\mathfrak{G}$ acting on $\Omega$ is an automorphism group for $\Xs$ if and only if for all $\xs \in \Omega$ and all $g \in \mathfrak{G}$ we have that $\pi(\xs) = \pi(\xs^g)$. 
\end{definition}

Note that we do not require the automorphism group to be maximal, that is, it can be a subgroup of a different automorphism group for the same set of random variables. Moreover, note that the definition of an automorphism group is independent of the particular representation of the probabilistic model. For particular representations, there are efficient algorithms for computing the automorphism groups exploiting the structure of relational and propositional graphical models~\cite{niepertorbits,niepert2012lmcmc,hai2012automorphism}.

\begin{figure}[t!]
\begin{center}
\includegraphics[width=0.32\textwidth]{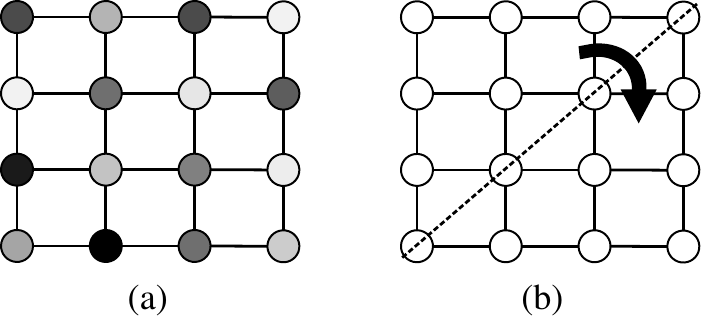}
\caption{A ferromagnetic Ising model with constant interaction strength. In the presence of an external field, that is, when the variables have different unary potentials, the probabilistic model is asymmetric (a). However, the model is rendered symmetric by assuming a constant external field (b). In this case, the symmetries of the model are generated by the reflection and rotation automorphisms.}
\label{fig:ising}
\end{center}
\end{figure} 

Most probabilistic models are asymmetric. For instance, the Ising model which is used in numerous applications, is asymmetric if we assume an external field as it leads to different unary potentials. However, we can make the model symmetric simply by assuming a constant external field. Figure~\ref{fig:ising} depicts this situation. The framework we propose in this paper will take advantage of such an over-symmetric model without biasing the probability estimates.

\subsection{Exploiting Symmetries for Lifted Inference}

The advent of high-level representations of probabilistic graphical models, such as plate models and relational representations~\citep{Getoor:2007,DeRaedt2008-PILP}, have motivated a new class of \emph{lifted inference} algorithms~\citep{Poole2003}. These algorithms exploit the high-level structure and symmetries to speed up inference~\citep{Kersting:2012}.
Surprisingly, they perform tractable inference even in the absence of conditional independencies.

Our current understanding of exact lifted inference is that syntactic properties of relational representations permit efficient lifted inference~\citep{VdBNIPS11,JaegerStarAI12,373041,gribkoff2014understanding}.
The Appendix will review such representations, and Markov logic in particular.
More recently, it has been shown that (partial) exchangeability as a statistical property can explain numerous results in this literature~\cite{DBLP:conf/aaai/NiepertB14}. Indeed, there are deep connections between automorphisms and exchangeability~\citep{niepertorbits,niepert2013symmetry,hai2012automorphism,BuiHB12}.
Moreover, the (fractional) automorphisms of the graphical model representation have been related to lifted inference and exploited for more efficient inference~\citep{niepertorbits,hai2012automorphism,noessner:2013,mladenov2013lifted}.
In particular, there are a number of sampling algorithms that take advantage of symmetries~\cite{venugopal2012lifting,gogate2012advances}. However, these approaches expect a relational representation and require the model to be symmetric. 

\subsection{Finite Markov Chains}

Given a finite set $\Omega$ a \emph{Markov chain} defines a random walk $(\mathbf{x}_0, \mathbf{x}_1, ...)$ on elements of $\Omega$ with the property that the conditional distribution of $\mathbf{x}_{n+1}$  given ($\mathbf{x}_0, \mathbf{x}_1, ..., \mathbf{x}_n)$ depends only on $\mathbf{x}_n$. For all $\xs, \ys \in \Omega$, $P(\xs \rightarrow \ys)$ is the chain's probability to transition from $\xs$ to $\ys$, and $P^t(\xs \rightarrow \ys)=P^{t}_{\xs}(\ys)$ the probability of being in state $\ys$ after $t$ steps if the chain starts at state $\xs$. We often refer to the conditional probability matrix $P$ as the \emph{kernel} of the Markov chain. A Markov chain is \emph{irreducible} if for all $\xs, \ys \in \Omega$ there exists a $t$ such that $P^t(\xs \rightarrow \ys) > 0$ and \emph{aperiodic} if for all $\xs \in \Omega$, $\mathsf{gcd}\{t \geq 1\ |\ P^t(\xs \rightarrow \xs) > 0\} = 1$. 


\begin{theorem}
Any irreducible and aperiodic Markov chain has exactly one stationary distribution.
\end{theorem}

A distribution $\pi$ on $\Omega$ is reversible for a Markov chain with state space $\Omega$ and transition probabilities $P$, if for every $\xs,\ys \in \Omega$
$$\pi(\xs)P(\xs \rightarrow \ys) = \pi(\ys)P(\ys \rightarrow \xs).$$
We say that a Markov chain is reversible if there exists a reversible distribution for it.
The AI literature often refers to reversible Markov chains as Markov chains satisfying the detailed balance property.

\begin{theorem}
Every reversible distribution for a Markov chain is also a stationary distribution for the chain.
\end{theorem}

\subsubsection{Markov Chains for Probability Estimation}

Numerous approximate inference algorithms for probabilistic graphical models draw sample points from a Markov chain whose stationary distribution is that of the probabilistic model, and use the sample points to estimate marginal probabilities. Sampling approaches of this kind are referred to as Markov chain Monte Carlo methods. We discuss the Gibbs sampler, a sampling algorithm often used in practice.

Let $\mathbf{X}$ be a finite set of random variables with  probability distribution $\pi$.
The Markov chain for the \emph{Gibbs sampler} is a Markov chain $\mathcal{M}=(\mathbf{x}_0, \mathbf{x}_1, ...)$ which, being in state $\mathbf{x}_t$, performs the following steps at time $t+1$:
\begin{enumerate}
\item Select a variable $X \in \mathbf{X}$ uniformly at random;
\item Sample $\mathbf{x}'_{t+1}(X)$, the value of $X$ in the state $\mathbf{x}'_{t+1}$, according to the conditional $\pi$-distribution of $X$ given that all other variables take their values according to $\mathbf{x}_t$; and
\item Let $\mathbf{x}'_{t+1}(Y)=\mathbf{x}_{t}(Y)$ for all variables $Y \in \mathbf{X} \setminus \{X\}$.
\end{enumerate}

The Gibbs chain is aperiodic and has $\pi$ as a stationary distribution. If the chain is irreducible, then the marginal estimates based on sample points drawn from the chain are  unbiased once the chain reaches the stationary distribution. 

Two or more Markov chains can be combined by constructing mixtures and compositions of the kernels~\cite{tierney1994}.
Let $P_1$ and $P_2$ be the kernels for two Markov chains $\mathcal{M}_1$ and $\mathcal{M}_2$ both with stationary distribution $\pi$. Given a positive probability $0 < \alpha < 1$, a \emph{mixture} of the Markov chains is a Markov chain where, in each iteration, kernel $P_1$ is applied with probability $\alpha$ and kernel $P_2$ with probability $1 - \alpha$. The resulting Markov chain has $\pi$ as a stationary distribution. 
The following result relates properties of the individual chains to properties of their mixture.

\begin{theorem}[\citenp{tierney1994}~1994]
A mixture of two Markov chains $\mathcal{M}_1$ and $\mathcal{M}_2$ is irreducible and aperiodic if at least one of the chains is irreducible and aperiodic. 
\end{theorem}

For a more in-depth discussion of combining Markov chains and the application to machine learning, we refer the interested reader to an overview paper~\cite{DBLP:journals/ml/AndrieuFDJ03}.


\section{Mixing Symmetric and Asymmetric\\ Markov Chains}

We propose a novel MCMC framework that constructs \emph{mixtures} of Markov chains where one of the chains operates on the \emph{approximate symmetries} of the probabilistic model. The framework assumes a base Markov chain $\mathcal{M}_{\mathtt{B}}$ such as the Gibbs chain, the MC-SAT chain~\cite{Poon:2006}, or any other MCMC algorithm. We then construct a mixture of the base chain and an Orbital Metropolis chain which exploits approximate symmetries for its proposal distribution. Before we describe the approach in more detail, let us first review Metropolis samplers.

\subsection{Metropolis-Hastings Chains}
The construction of a Metropolis-Hastings Markov chain is a popular general procedure for designing reversible Markov chains for MCMC-based estimation of marginal probabilities. Metropolis-Hastings chains are associated with a proposal distribution $Q(\cdot | \xs)$ that is utilized to \emph{propose} a move to the next state given the current state $\mathbf{x}$. The closer the proposal distribution to the distribution $\pi$ to be estimated, that is, the closer $Q(\xs \mid \mathbf{x}_t)$ to $\pi(\xs)$ for large $t$,  the better the convergence properties of the Metropolis-Hastings chain. 

We first describe the Metropolis algorithm, a special case of the Metropolis-Hastings algorithm~\cite{mcmcbook}. Let $\mathbf{X}$ be a finite set of random variables with  probability distribution $\pi$ and let $\Omega$ be the set of states of the random variables. The Metropolis chain is governed by a transition graph $G = (\Omega, \mathbf{E})$ whose nodes correspond to states of the random variables. Let $\mathtt{n}(\mathbf{x})$ be the set of neighbors of state  $\mathbf{x}$ in $G$, that is, all states reachable from $\mathbf{x}$ with a single transition. The Metropolis chain with graph $G$ and distribution $\pi$ has transition probabilities\\

\noindent$P(\mathbf{x} \rightarrow \mathbf{y})=$
\begin{equation*}
   \begin{cases}
    \frac{1}{|\mathtt{n}(\mathbf{x})|}\min\left\{\frac{\pi(\mathbf{y})|\mathtt{n}(\mathbf{x})|}{\pi(\mathbf{x})|\mathtt{n}(\mathbf{y})|}, 1\right\}, &\hspace{-20mm} \text{if $x$ and $y$ are neighbors}\\
    0, & \hspace{-20mm} \text{if $x \neq y$ are not neighbors}\\
    1 - \sum\limits_{y' \in \mathtt{n}(\mathbf{x})}\frac{1}{|\mathtt{n}(\mathbf{x})|}\min\left\{\frac{\pi(\mathbf{y'})|\mathtt{n}(\mathbf{x})|}{\pi(\mathbf{x})|\mathtt{n}(\mathbf{y'})|}, 1\right\}, & \text{if $x$ = $y$}.
  \end{cases}
\end{equation*}

Being in state $\mathbf{x}_t$ of the Markov chain $\mathcal{M}=(\mathbf{x}_0, \mathbf{x}_1, ...)$, the Metropolis sampler therefore performs the following steps at time $t+1$:

\begin{enumerate}
\item Select a state $\ys$ from $\mathtt{n}(\mathbf{x}_t)$, the neighbors of $\xs_t$, uniformly at random;
\item Let $\mathbf{x}_{t+1}=\mathbf{y}$ with probability $\min\left\{\frac{\pi(\mathbf{y})|\mathtt{n}(\mathbf{x})|}{\pi(\mathbf{x})|\mathtt{n}(\mathbf{y})|}, 1\right\}$;
\item Otherwise, let $\mathbf{x}_{t+1}=\xs_t$.
\end{enumerate}

Note that the proposal distribution $Q( \cdot | \xs)$ is simply the uniform distribution on the set of $\xs$'s neighbors. 
It is straight-forward to show that $\pi$ is a stationary distribution for the Metropolis chain  by showing that $\pi$ is a reversible distribution for it~\cite{mcmcbook}.

Now, the performance of the Metropolis chain hinges on the structure of the graph $G$. We would like the graph structure to facilitate global moves between high probability modes, as opposed to the local moves typically performed by MCMC chains. To design such a graph structure, we take advantage of approximate symmetries in the model. 

\subsection{Orbital Metropolis Chains}

We propose a novel class of \emph{orbital} Metropolis chains that move between approximate symmetries of a distribution. The approximate symmetries form an automorphism group $\mathfrak{G}$.
We will discuss approaches to obtain such an automorphism group in Section~\ref{s:symmetrization}.
Here, we introduce a novel Markov chain that takes advantage of the approximate symmetries.

Given a distribution $\pi$ over random variables $\Xs$ with state space $\Omega$, and a permutation group $\mathfrak{G}$ acting on $\Omega$, the orbital Metropolis chain $\mathcal{M}_{\mathtt{S}}$ for $\mathfrak{G}$ performs the following steps:

\begin{enumerate}
\item Select a state $\ys$ from $\mathbf{x}_t^{\mathfrak{G}}$, the orbit of $\xs_t$, uniformly at random;
\item Let $\mathbf{x}_{t+1}=\mathbf{y}$ with probability $\min\left\{\frac{\pi(\mathbf{y})}{\pi(\mathbf{x})}, 1\right\}$;
\item Otherwise, let $\mathbf{x}_{t+1}=\xs_t$.
\end{enumerate}

Note that a permutation group acting on $\Omega$ partitions the states into disjoint orbits. The orbital Metropolis chain simply moves between states in the same orbit.
Hence, two states in the same orbit have the same number of neighbors and, thus, the expressions  cancel out in line 2 above. It is straight-forward to show that the chain  $\mathcal{M}_{\mathtt{S}}$ is reversible and, hence, that it has $\pi$ as a stationary distribution. However, the chain is \emph{not} irreducible as it never moves between states that are not symmetric with respect to the permutation group $\mathfrak{G}$. In the binary case, for example, it cannot reach states with a different Hamming weight from the initial state.

\subsection{Lifted Metropolis-Hastings}

To obtain an irreducible Markov chain that exploits approximate symmetries, we construct a mixture of (a) some base chain $\mathcal{M}_{\mathtt{B}}$ with stationary distribution $\pi$ for which we know that it is irreducible and aperiodic; and (b) an orbital Metropolis chain $\mathcal{M}_{\mathtt{S}}$.
We can prove the following theorem.

\begin{theorem}
Let $\Xs$ be a set of random variables with distribution $\pi$ and approximate automorphisms $\mathfrak{G}$. Moreover, let $\mathcal{M}_{\mathtt{B}}$ be an aperiodic and irreducible Markov chain with stationary distribution $\pi$, and let $\mathcal{M}_{\mathtt{S}}$ be the orbital Metropolis chain for $\Xs$ and $\mathfrak{G}$. The mixture of $\mathcal{M}_{\mathtt{B}}$ and $\mathcal{M}_{\mathtt{S}}$ is aperiodic, irreducible, and has $\pi$ as its unique stationary distribution.
\end{theorem}

\begin{figure}[t!]
\begin{center}
\includegraphics[width=0.43\textwidth]{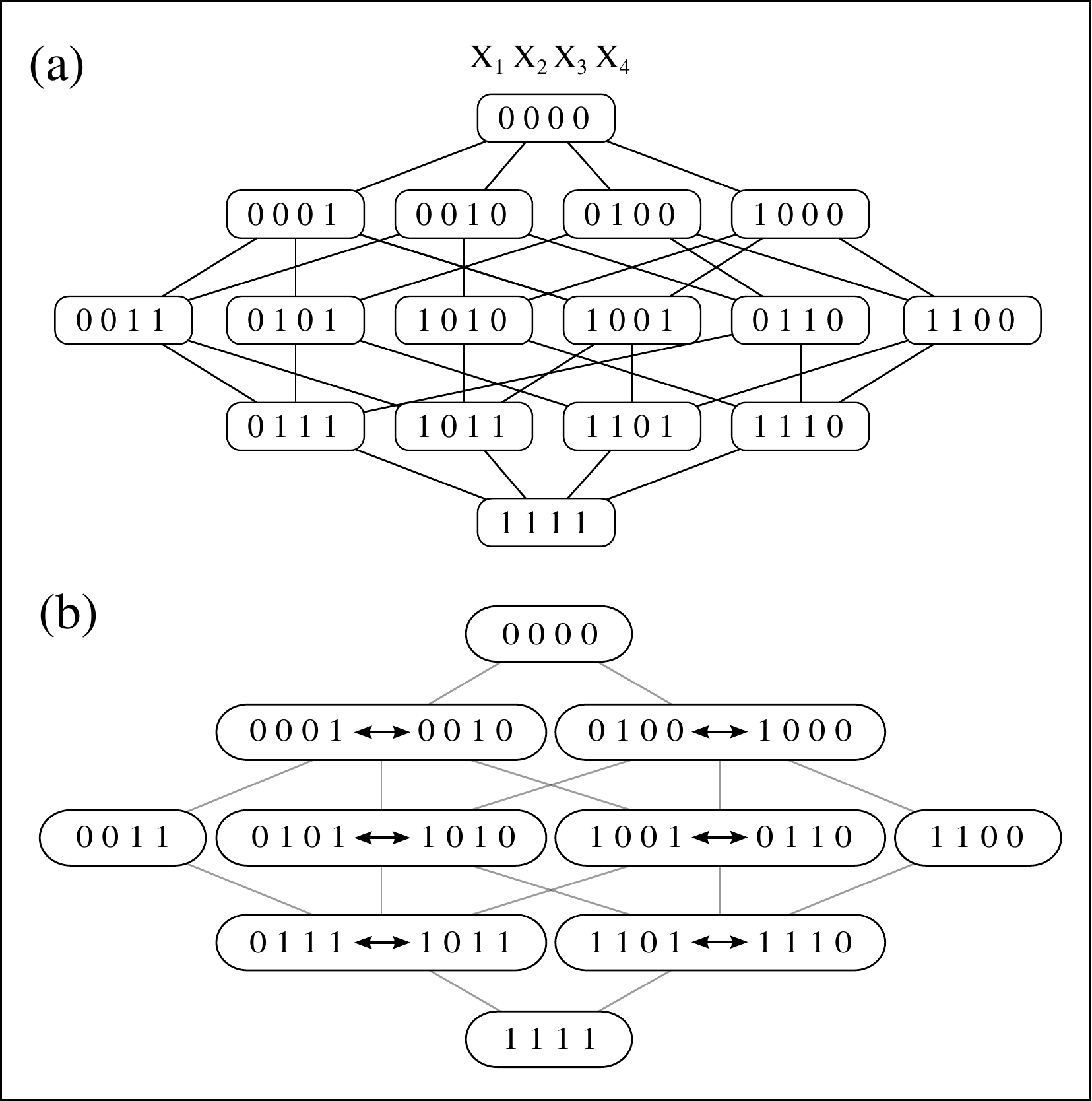}
\caption{The state space (self-arcs are omitted) of (a)~the Gibbs chain for four binary random variables and (b)~the orbit partition of its state space  induced by the permutation group generated by the permutation ($X_1\ X_2)(X_3\ X_4)$. The permutations are approximate symmetries, derived from an over-symmetric approximation of the original model. The Gibbs chain proposes moves to states whose Hamming distance to the current state is at most $1$. The orbital Metropolis chain, on the other hand, proposes moves between orbit elements which have a Hamming distance of up to $4$. The mixture of the two chains leads to faster convergence while maintaining an unbiased stationary distribution.}
\label{fig:lumping}
\end{center}
\end{figure} 

The mixture of the base chain and the orbital Metropolis chain has several advantages. First, it exploits the approximate symmetries of the model which was shown to be advantageous for marginal probability estimation~\cite{conf/nips/BroeckD13}. Second, the mixture of Markov chains performs wide ranging moves via the orbital Metropolis chain, exploring the state space more efficiently and, therefore, improving the quality of the probability estimates. Figure~\ref{fig:lumping} depicts the state space and the transition graph of (a)~the Gibbs chain and (b)~the mixture of the Gibbs chain and an orbital Metropolis chain. It illustrates that the mixture is able to more freely move about the state space by jumping between orbit states. For instance, moving from state $0110$ to $1001$ would require $4$ steps of the Gibbs chain but is possible in one step with the mixture of chains. The larger the size of the automorphism groups, the more densely connected is the transition graph. 
Since the moves of the orbital Metropolis chain are between approximately symmetric states of the random variables, it does not suffer from the problem of most proposals being rejected. We will be able to verify this hypothesis empirically. 

The general Lifted Metropolis-Hastings framework can be summarized as follows.
\begin{enumerate}
\item Obtain an approximate automorphism group $\mathfrak{G}$;
\item Run the following mixture of Markov chains:
\begin{enumerate}
\item With probability $0 < \alpha < 1$, apply the kernel of the base chain $\mathcal{M}_{\mathtt{B}}$;
\item Otherwise, apply the kernel of the orbital Metropolis chain $\mathcal{M}_{\mathtt{S}}$ for $\mathfrak{G}$.
\end{enumerate}
\end{enumerate}

Note that the proposed approach is a generalization of lifted MCMC~\cite{niepert2013symmetry,niepertorbits}, essentially using it as a subroutine, and that all MH proposals are accepted if $\mathfrak{G}$ is an automorphism group of the original model.
Moreover, note that the framework allows one to combine multiple orbital Metropolis chains with a base chain.

\section{Approximate Symmetries} \label{s:symmetrization}

The Lifted Metropolis-Hastings algorithm assumes that a permutation group $\mathfrak{G}$ is given, representing the approximate symmetries.
We now discuss several approaches to the computation of such an automorphism group.
While it is not possible to go into technical detail here, we will provide pointers to the relevant literature.

There exist several techniques to compute the \emph{exact symmetries} of a graphical model and construct~$\mathfrak{G}$; see~\citep{niepertorbits,hai2012automorphism}.
The color refinement algorithm is also well-studied in lifted inference~\cite{DBLP:conf/aaai/KerstingMGG14}. It can find (exact) orbits of random variables for a slightly weaker notion of symmetry, called fractional automorphism. These techniques all require some form of exact symmetry to be present in the model.

Detecting \emph{approximate symmetries} is a problem that is largely open.
One key idea is that of an \emph{over-symmetric approximations}~(OSAs)~\cite{conf/nips/BroeckD13}. 
Such approximations are derived from the original model by rendering the model more symmetric.  After the computation of an over-symmetric model, we can apply existing tools for exact symmetry detection.
Indeed, the exact symmetries of an approximate model are approximate symmetries of the exact model.
These symmetrization techniques are indispensable to our algorithm.

\subsubsection{Relational Symmetrization}

Existing symmetrization techniques operate on relational representations, such as Markov logic networks (MLNs). The full paper reviews MLNs and shows a web page classification model. Relational models have numerous symmetries. For example, swapping the web pages $\pagea$ and $\pageb$ does not change the MLN. This permutation of constants induces a permutations of random variables (e.g., between $\pageclass(\pagea,\Faculty)$ and $\pageclass(\pageb,\Faculty)$).
Unfortunately, hard and soft evidence breaks symmetries, even in highly symmetric relational models~\cite{conf/nips/BroeckD13}. When the variables $\pageclass(\pagea,\Faculty)$ and $\pageclass(\pageb,\Faculty)$ get assigned distinct soft evidence, the symmetry between $\pagea$ and $\pageb$ is removed, and lifted inference breaks down.\footnote{Solutions to this problem exist if the soft evidence is on a single unary relation~\cite{BuiHB12}}
Similarly, when the $\linked$ relation is given, its graph is unlikely to be symmetric~\cite{erdHos1963asymmetric}, which in turn breaks the symmetries in the MLN.
These observations motivated research on OSAs. \citet{conf/nips/BroeckD13} propose to approximate binary relations, such as $\linked$, by a low-rank Boolean matrix factorization. 
\citet{DBLP:conf/pkdd/VenugopalG14} cluster the constants in the domain of the MLN. \citet{singla2014approximate} present a message-passing approach to clustering similar constants.

\subsubsection{Propositional Symmetrization}

A key property of our LMH algorithm 
is that it operates at the propositional level, regardless of how the graphical model was generated. 
It also means that the relational symmetrization approaches outlined above are inadequate in the general case. Unfortunately, we are not aware of any work on OSAs of propositional graphical models.
However, some existing techniques provide a promising direction. First, basic clustering can group together similar potentials.  Second, the low-rank Boolean matrix factorization used for relational approximations can be applied to any graph structure, including graphical models. Third, color passing techniques for exact symmetries operate on propositional models~\cite{DBLP:conf/uai/KerstingAN09,DBLP:conf/aaai/KerstingMGG14}. Combined with early stopping, they can output  approximate variable orbits.

\subsubsection{From OSAs to Automorphisms}

Given an OSA of our model, we need to compute an automorphism group $\mathfrak{G}$ from it. 
The obvious choice is to compute the exact automorphisms from the OSA. While this works in principle, it may not be optimal.
Let us first consider the following two concepts.
When a group $\mathfrak{G}$ operates on a set $\Omega$, only a subset of the elements in $\Omega$ can actually be mapped to an element other than itself. When $\Omega$ is the set of random variables, we call these elements the \emph{moved variables}. When $\Omega$ is the set of potentials in a probabilistic graphical model, we call these the \emph{moved potentials}.
It is clear that we want $\mathfrak{G}$ to move many random variables, as this will create the largest jumps and improve the mixing behavior. However, each LMH step comes at a cost: in the second step of the algorithm, the probability of the proposed approximately-symmetric state $\pi(\mathbf{y})$ is estimated. This requires the re-evaluation of all potentials that are moved by~$\mathfrak{G}$. Thus, the time complexity of an orbital Metropolis step is linear in the number of moved potentials. It will therefore be beneficial to construct \emph{subgroups} of the automorphism group of the OSA and, in particular, ones that move many variables and few potentials. The full paper discusses a heuristic to construct such subgroups.

\section{Experiments}

The LMH algorithm is implemented in the GAP algebra system which provides basic algorithms for automorphism groups such as the product replacement algorithm that allows one to sample uniformly from orbits of states~\cite{niepertorbits}.

For our first experiments, we use the standard WebKB data set, consisting of web pages from four computer science departments~\cite{craven&slattery01}. The data has information about approximately 800 words that appear on 1000 pages, 7 page labels and links between web pages. There are 4 folds, one for each university.
We use the standard MLN structure for the WebKB domain, which has MLN formulas of the form shown above, but for all combinations of labels and words, adding up to around 5500 first-order MLN formulas. We learn the MLN parameters using Alchemy.

\begin{figure}[t!]
  \centering
  \subfigure[Texas - Iterations]{%
    \includegraphics[height=\plotheight]{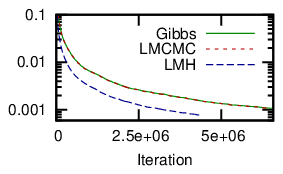}
  }
  \subfigure[Texas - Time]{%
    \includegraphics[height=\plotheight]{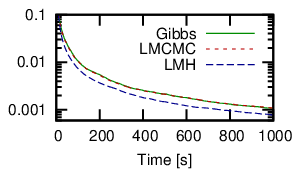}
  }
  \subfigure[Washington - Iterations]{%
    \includegraphics[height=\plotheight]{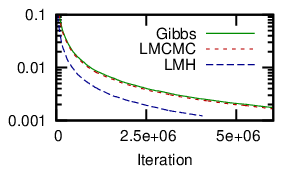}
  }
  \subfigure[Washington - Time]{%
    \includegraphics[height=\plotheight]{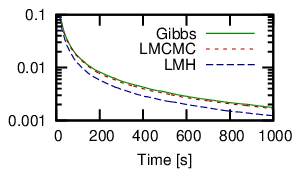}
  }
  \caption{WebKB - KL Divergence of Texas and Washington} \label{fig:webkb}
\end{figure}

\begin{figure}[htb]
  \centering
  \subfigure[Cornell - Iterations]{%
    \includegraphics[height=\plotheight]{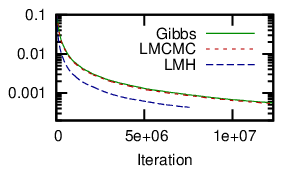}
  }
  \subfigure[Cornell - Time]{%
    \includegraphics[height=\plotheight]{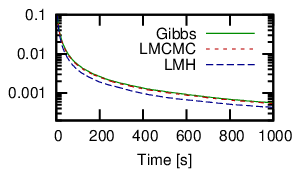}
  }
  \subfigure[Wisconsin - Iterations]{%
    \includegraphics[height=\plotheight]{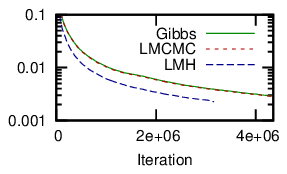}
  }
  \subfigure[Wisconsin - Time]{%
    \includegraphics[height=\plotheight]{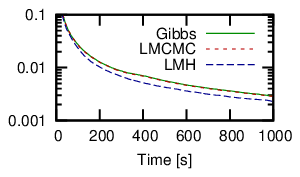}
  }
  \caption{WebKB - KL Divergence of Cornell and Wisconsin \label{fig:webkbextra}}
\end{figure}

We consider a collective classification setting, where we are given the link structure and the word content of each web page, and want to predict the page labels. We run Gibbs sampling and the Lifted MCMC algorithm~\cite{niepertorbits}, and show the average KL divergence between the estimated and true marginals in Figures~\ref{fig:webkb} and~\ref{fig:webkbextra}.  When true marginals are not computable, we used a very long run of a Gibbs sampler for the gold standard marginals.  Since every web page contains a unique set of words, the evidence on the word content creates distinct soft evidence on the page labels. Moreover, the link structure is largely asymmetric and, therefore, there are no exploitable exact symmetries and Lifted MCMC coincides with Gibbs sampling.
Next we construct an OSA using a rank-5 approximation of the link structure~\cite{conf/nips/BroeckD13} and group the potential weights into 6 clusters. From this OSA we construct a set of automorphisms that is efficient for LMH (see Appendix~\ref{s:heur}). Figures~\ref{fig:webkb} and~\ref{fig:webkbextra} show that the LMH chain, with mixing parameter $\alpha=4/5$,  has a lower KL divergence than Gibbs and Lifted MCMC vs.~the number of iterations. Note that there is a slight overhead to LMH because the orbital Metropolis chain is run between base chain steps. Despite this overhead, LMH outperforms the baselines as a function of time. The orbital Metropolis chain accepts approximately 70\% of its proposals.

\begin{figure}[t!]
  \centering
  \includegraphics[width=0.8\columnwidth]{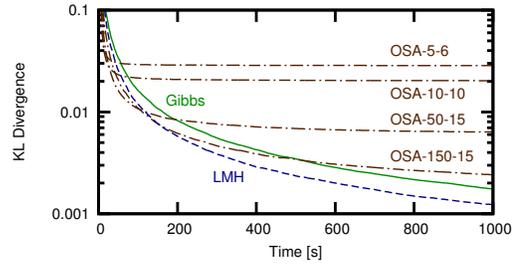}
  \caption{LMH vs.\ over-symmetric approximations (OSA) on WebKB Washington. OSA-$r$-$c$ denotes binary evidence of Boolean rank $r$ and $c$ clusters of formula weights.} \label{fig:osa}
\end{figure}

Figure~\ref{fig:osa} illustrates the effect of running Lifted MCMC on OSA, which is the current state-of-the-art approach for asymmetric models. As expected, the drawn sample points produce biased estimates. As the quality of the approximation increases, the bias reduces, but so do the speedups. LMH does not suffer from a bias. Moreover, we observe that its performance is stable across different OSAs (not depicted).

\begin{figure}[t!]
  \centering
  \subfigure[Ising - Iterations]{%
    \includegraphics[height=\plotheight]{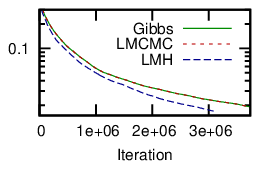}
  }~~~~~~
  \subfigure[Ising - Time]{%
    \includegraphics[height=\plotheight]{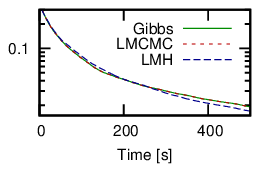}
  } \\
  \subfigure[Chimera - Iterations]{%
    \includegraphics[height=\plotheight]{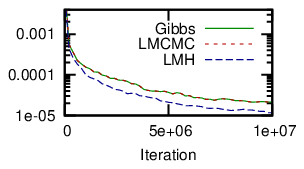}
  }
  \subfigure[Chimera - Time]{%
    \includegraphics[height=\plotheight]{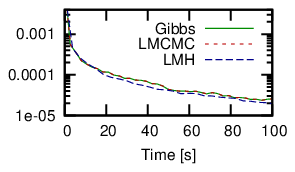}
  }
  \caption{\label{props}KL Divergences for the propositional models.}
\end{figure}


We also ran experiments for two propositional models that are frequently used in real world applications. The first model is a $100$x$100$ ferromagnetic Ising model with constant interaction strength and external field (see Figure~\ref{fig:ising}(a) for a $4$x$4$ version). Due to the different potentials induced by the external field, the model has no symmetries. We use the model without external field to compute the approximate symmetries. The automorphism group representing these symmetries is generated by the rotational and reflectional symmetries of the grid model (see Figure~\ref{fig:ising}(b)). As in the experiments with the relational models, we used the mixing parameter $\alpha = 4/5$ for the LMH algorithm. Figure~\ref{props}(c) and (d) depicts the plots of the experimental results. The LMH algorithm performs better with respect to the number of iterations and, to a lesser extent, with respect to time. 

We also ran experiments on the Chimera model which has recently received some attention as it was used to assess the performance of quantum annealing~\cite{boixo:2013}. We used exactly the model as described in~\citet{boixo:2013}. This model is also asymmetric but can be made symmetric by assuming that all pairwise interactions are identical. The KL divergence vs. number of iterations and  vs. time in seconds is plotted in Figure~\ref{props}(a) and (b), respectively. Similar to the results for the Ising model, LMH outperforms Gibbs and LMCMC both with respect to the number of iterations and wall clock time. 
In summary, the LMH algorithm outperforms standard sampling algorithms on these propositional models in the absence of any symmetries. We used very simple symmetrization strategies for the experiments. This demonstrates that the LMH framework is powerful and allows one to design state-of-the-art sampling algorithms.

\section{Conclusions}

We have presented a Lifted Metropolis-Hastings algorithms capable of mixing two types of Markov chains. The first is a non-lifted base chain, and the second is an orbital Metropolis chain that moves between approximately symmetric states. 
This allows lifted inference techniques to be applied to asymmetric graphical models.

\subsubsection{ Acknowledgments}
This work was supported by 
the Research Foundation-Flanders (FWO-Vlaanderen).

%


\bibliographystyle{aaai}
\bibliography{paper}



\appendix

\section{Markov Logic Networks} \label{s:mlns}

We first introduce some standard concepts from first-order logic.
An \textit{atom} $\P(t_1, \dots , t_n)$ consists of a predicate $\P/n$ of 
arity $n$ followed by $n$ argument terms $t_i$, which are either \textit{constants}, $\{\pagea,\pageb,\dots\}$ or \textit{logical variables} $\{\xl, \yl, \dots\}$.
A formula combines atoms with connectives (e.g., $\land$, $\Leftrightarrow$).
A formula is {\em ground} if it contains no logical variables. The groundings of a formula are obtained by instantiating the variables with particular constants.

Many statistical relational languages have been proposed in recent years~\citep{Getoor:2007,DeRaedt2008-PILP}. One such language is \emph{Markov logic networks}~(MLN)~\citep{richardson2006markov}.
An MLN is a set of tuples $(w,f)$, where $w$ is a real number representing a weight and $f$ is a formula in first-order logic. 
Consider for example the MLN
\small
\mln{ \begin{align*}
1.3 \quad & \pageclass(\xl,\Faculty) \Rightarrow \has(\xl,\Hours)\\
1.5 \quad & \pageclass(\xl,\Faculty) \land \linked(\xl,\yl) \Rightarrow \pageclass(\yl,\Course)
\end{align*}}
\normalsize
which states that faculty web pages are more likely to contain the word ``hours'', and that faculty pages are more likely to link to course pages.

Given a domain of constants $\dom$, a first-order MLN \(\Delta\) induces a {\em grounding}, which is the MLN obtained by replacing each formula in \(\Delta\) with all its groundings.
For the domain $\dom = \{\pagea,\pageb\}$ (i.e., two pages), the MLN represents the following grounding.
\small
\mln{
  \begin{align*}
1.3 \quad & \pageclass(\pagea,\Faculty) \Rightarrow \has(\pagea,\Hours)\\
1.3 \quad & \pageclass(\pageb,\Faculty) \Rightarrow \has(\pageb,\Hours)\\
1.5 \quad & \pageclass(\pagea,\Faculty) \land \linked(\pagea,\pageb) \Rightarrow \pageclass(\pageb,\Course)\\
1.5 \quad & \pageclass(\pageb,\Faculty) \land \linked(\pageb,\pagea) \Rightarrow \pageclass(\pagea,\Course)\\
1.5 \quad & \pageclass(\pagea,\Faculty) \land \linked(\pagea,\pagea) \Rightarrow \pageclass(\pagea,\Course)\\
1.5 \quad & \pageclass(\pageb,\Faculty) \land \linked(\pageb,\pageb) \Rightarrow \pageclass(\pageb,\Course)
  \end{align*}
}
\normalsize
This grounding has ten random variables, yielding a distribution over \(2^{10}\) possible worlds. The weight
of each world is the product of the expressions \(\exp(w)\), where \((w,f)\) is a ground MLN formula and \(f\) is satisfied by the world.

\section{Approximate Automorphism Heuristic} \label{s:heur}

Given an OSA, we construct a set of approximate automorphisms as follows.
First, we compute the exact automorphisms $\mathfrak{G}_1$ of the OSA. 
Second, we compute the variable orbits of $\mathfrak{G}_1$, grouping together all variables that can be mapped into each other.
Then, for every orbit $O$, we construct a set of automorphisms as follows.
We greedily search for a $O' \subseteq O$ such that the symmetric group $\mathfrak{G}_{O'}$ on $O'$ maximizes the ratio between the number of moved variables (i.e., $|O'|$) and the number of moved potentials, while keeping the number of moved potentials bounded by a constant $K$. This guarantees that $\mathfrak{G}_{O'}$ yields an efficient orbital Metropolis chain. Finally, we remove $O'$ from $O$ and recurse until $O$ is empty.
From this set of symmetric groups $\mathfrak{G}_{O'}$, we construct a set of orbital Metropolis chains, each with it own set of moved potentials.

\end{document}